\documentclass[11pt]{article}
\usepackage{acl2015}
\usepackage{times}
\usepackage{url}
\usepackage{latexsym}
\usepackage{graphicx,amsmath,amssymb,booktabs,color,bm,amsthm}
\usepackage{enumitem}
\usepackage{multirow}
\usepackage[font=small,labelfont=bf]{caption}
\usepackage{todonotes}
\newcommand{\Note}[2]{}
\newcommand{\SideNote}[2]{}
\renewcommand{\Note}[2]{\todo[color=#1,size=\small, inline=true]
{#2}} 
\renewcommand{\SideNote}[2]{\todo[color=#1,size=\small]{#2}}

\title{Language Models for Image Captioning: The Quirks and What Works}

\author{Jacob Devlin$^\bigstar$, Hao Cheng{\small $^\spadesuit$}, Hao Fang{\small $^\spadesuit$}, Saurabh Gupta{\small $^\clubsuit$}, \\ {\bf Li Deng, Xiaodong He$^\bigstar$, Geoffrey Zweig$^\bigstar$, Margaret Mitchell$^\bigstar$} \\
\\
{\bf Microsoft Research} \\
\\
{\small $\bigstar$} Corresponding authors: \{jdevlin,xiaohe,gzweig,memitc\}@microsoft.com \\
{\small $\spadesuit$} University of Washington \\
{\small $\clubsuit$} University of California at Berkeley
}

\date{}

\begin{document}
\maketitle

\begin{abstract}
Two recent approaches have achieved state-of-the-art results in image captioning. The first uses a pipelined process where a set of candidate words is generated by a convolutional neural network (CNN) trained on images, and then a maximum entropy (ME) language model is used to arrange these words into a coherent sentence. The second uses the penultimate activation layer of the CNN as input to a recurrent neural network (RNN) that then generates the caption sequence. In this paper, we compare the merits of these different language modeling approaches for the first time by using the same state-of-the-art CNN as input. We examine issues in the different approaches, 
including linguistic irregularities, caption repetition, and data set overlap.  By combining key aspects of the ME and RNN methods, we
achieve a new record performance over previously published results on the benchmark COCO dataset.  However, the gains we see in BLEU do not translate to human judgments.
\end{abstract}

\section{Introduction}
\label{sec:intro}
Recent progress in automatic image captioning has shown that an image-conditioned
language model can be very effective at generating captions.
Two leading approaches have been explored for this task.  The first decomposes the problem into an initial step that uses a convolutional neural network to predict a bag of words that are likely to be present in a caption; then in a second step, a maximum entropy language model (ME LM) is used to generate a sentence that covers a minimum number of the detected words
\cite{MSRCapVisCVPR2015}.  The second approach uses the activations from final hidden layer of an object detection CNN as the input to a recurrent neural network language model (RNN LM). This is referred to as a Multimodal Recurrent Neural Network (MRNN) \cite{Karpathy2015,Mao2015,Chen2015}. Similar in spirit is the the log-bilinear (LBL) LM of \newcite{Kiros2014LB}.

In this paper, we study the relative merits of these approaches. By using an identical state-of-the-art CNN as the input to RNN-based and ME-based models, we are able to empirically compare the strengths and weaknesses of the language modeling components.  We find that the approach of directly generating the text with an MRNN\footnote{In our case, a gated recurrent neural network (GRNN) is used \cite{Cho2014Arxiv}, similar to an LSTM.} outperforms the ME LM when measured by BLEU on the COCO dataset \cite{COCO2014},\footnote{This is the largest image captioning dataset to date.} but this recurrent model tends to reproduce captions in the training set. In fact, a simple $k$-nearest neighbor approach, which is common in earlier related work \cite{farhadi2010,mason2014}, performs similarly to the MRNN.  In contrast, the ME LM generates the most novel captions, and does the best at captioning images for which there is no close match in the training data. With a Deep Multimodal Similarity Model (DMSM) incorporated,\footnote{As described by  \newcite{MSRCapVisCVPR2015}.} the ME LM significantly outperforms other methods according to human judgments.  In sum, the contributions of this paper are as follows:
\vspace{-1ex}
\begin{enumerate}
\itemsep -1.5pt
\item We compare the use of discrete detections and continuous valued CNN activations as the conditioning information for language models trained to generate image captions.
\item We show that a simple $k$-nearest neighbor retrieval method performs at near state-of-the-art for this task and dataset. 
\item We demonstrate that a state-of-the-art MRNN-based approach tends to reconstruct previously seen captions; in contrast, the two stage ME LM approach achieves similar or better performance while generating relatively novel captions.
\item We advance the state-of-the-art BLEU scores on the COCO dataset.
\item We present human evaluation results on the systems with the best performance as measured by automatic metrics.
\item We explore several issues with the statistical models and the underlying COCO dataset, including linguistic irregularities, caption repetition, and data set overlap.
\end{enumerate}

\section{Models}
\label{sec:model}

All language models compared here are trained using output from the same state-of-the-art CNN. The CNN used is the 16-layer variant of VGGNet \cite{Simonyan2014} which was initially trained for the ILSVRC2014 classification task \cite{ILSVRC15}, and then fine-tuned on the Microsoft COCO data set  \cite{MSRCapVisCVPR2015,COCO2014}.

\subsection{Detector Conditioned Models}
We study the effect of leveraging an explicit detection step to find key objects/attributes in images before generation, examining both an ME LM approach as reported in previous work \cite{MSRCapVisCVPR2015}, and a novel LSTM approach  introduced here. Both use a CNN trained to output a bag of words indicating the words that are likely to appear in a caption, and both use a beam search to find a top-scoring sentence that contains a subset of the words. This set of words is dynamically adjusted to remove words as they are mentioned. 

We refer the reader to \newcite{MSRCapVisCVPR2015} for a full description of their ME LM approach, whose 500-best outputs we analyze here.\footnote{We will refer to this system as D-ME.}  We also include the output from their ME LM that leverages scores from a Deep Multimodal Similarity Model (DMSM) during $n$-best re-ranking. Briefly, the DMSM is a non-generative neural network model which projects both the image pixels and caption text into a comparable vector space, and scores their similarity.

In the LSTM approach, similar to the ME LM approach, we maintain a set of likely words $\mathcal{D}$ that have not yet been mentioned in the caption under construction. This set is initialized to all the words predicted by the CNN above some threshold $\alpha$.\footnote{In all experiments in this paper, $\alpha$=0.5.} The words already mentioned in the sentence history $h$ are then removed to produce a set of conditioning words $\mathcal{D} \setminus \{ {\bf h} \}$.  We incorporate this information within the LSTM by adding an additional input encoded to represent the remaining visual attributes $\mathcal{D} \setminus \{ {\bf h} \}$ as a continuous valued auxiliary feature vector \cite{mikolov2012context}.  This is encoded as
$f ({\bf s}_{h_{-1}} + \sum_{v \in \mathcal{D} \setminus \{ {\bf h} \}} {\bf g}_v
+ {\bf U} {\bf q}_{{\bf h}, \mathcal{D}})
$, 
where ${\bf s}_{h_{-1}}$ and ${\bf g}_v$ are respectively the continuous-space representations for last word $h_{-1}$ and detector $v \in \mathcal{D} \setminus \{ {\bf h} \}$, ${\bf U}$ is learned matrix for recurrent histories, and $f(\cdot)$ is the sigmoid transformation.

\label{ssec:rnn}

\subsection{Multimodal Recurrent Neural Network}
In this section, we explore a model directly conditioned on the CNN activations rather than a set of word detections. Our implementation is very similar to captioning models described in \newcite{Karpathy2015}, \newcite{Vinyals2015}, \newcite{Mao2015}, and \newcite{Donahue2014Arxiv}. This joint vision-language RNN is referred to as a Multimodal Recurrent Neural Network (MRNN).

In this model, we feed each image into our CNN and retrieve the 4096-dimensional final hidden layer, denoted as {\tt fc7}. The {\tt fc7} vector is then fed into a hidden layer $H$ to obtain a 500-dimensional representation that serves as the initial hidden state to a gated recurrent neural network (GRNN) \cite{Cho2014Arxiv}. The GRNN is trained jointly with $H$ to produce the caption one word at a time, conditioned on the previous word and the previous recurrent state. For decoding, we perform a beam search of size 10 to emit tokens until an END token is produced. We use a 500-dimensional GRNN hidden layer and 200-dimensional word embeddings. \label{ssec:clstm}

\subsection{$k$-Nearest Neighbor Model}
Both \newcite{Donahue2015} and \newcite{Karpathy2015} present a 1-nearest neighbor baseline. As a first step, we replicated these results using the cosine similarity of the {\tt fc7} layer between each test set image $t$ and training image $r$. We randomly emit one caption from $t$'s most similar training image as the caption of $t$.  As reported in previous results, performance is quite poor, with a BLEU score of 11.2\%.

However, we explore the idea that we may be able to find an optimal  $k$-nearest neighbor \textit{consensus caption}. We first select the $k=90$ nearest training images of a test image $t$ as above. We denote the union of training captions in this set as $C = c_1,...,c_{5k}$.\footnote{Each training image has 5 captions.} For each caption $c_i$, we compute the n-gram overlap F-score between $c_i$ and each other caption in $C$. We define the consensus caption $c^*$ to be caption with the highest mean n-gram overlap with the other captions in $C$. We have found it is better to only compute this average among $c_i$'s $m=125$  most similar captions, rather than all of $C$. The hyperparameters $k$ and $m$ were obtained by a grid search on the validation set.

\begin{figure}[t]
  \centering
  \includegraphics[width=\linewidth]{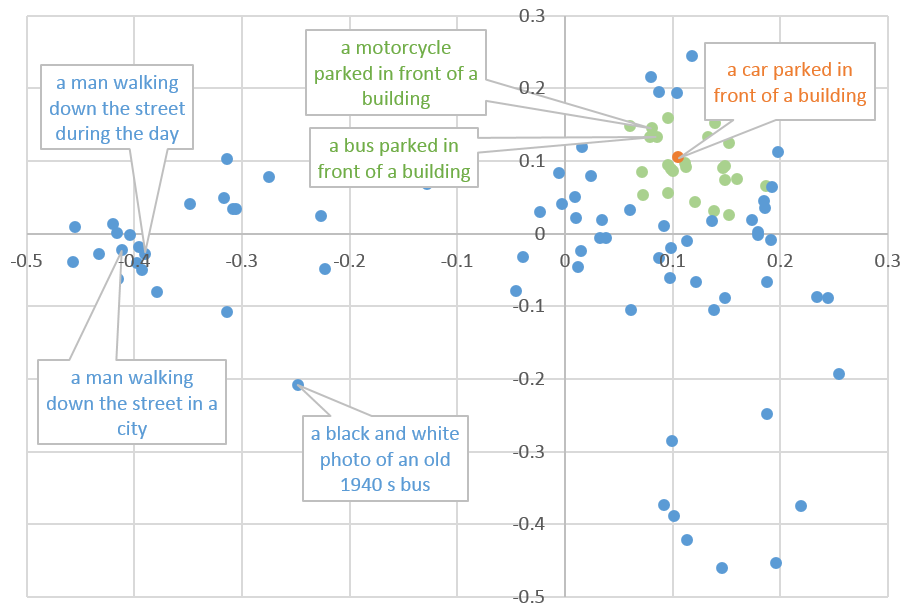}\\
  \caption{Example of the set of candidate captions for an image, the highest scoring $m$ captions (green) and the consensus caption (orange). This is a real example visualized in two dimensions.}\label{fig:consensus}
\end{figure}

A visual example of the consensus caption is given in Figure~\ref{fig:consensus}. Intuitively, we are choosing a single caption that may describe \textit{many} different images that are similar to $t$, rather than a caption that describes the \textit{single} image that is most similar to $t$. We believe that this is a reasonable approach to take for a retrieval-based method for captioning, as it helps ensure incorrect information is not mentioned.  Further details on retrieval-based methods are available in, e.g.,  \cite{Ordonez2011,Hodosh2013}.

\label{ssec:nearestneighbor}

\section{Experimental Results}
\label{sec:exp}
\subsection{The Microsoft COCO Dataset}
We work with the Microsoft COCO dataset \cite{COCO2014}, with 82,783 training images, and the validation set split into 20,243 validation images and 20,244 testval images.
Most images contain multiple objects and significant contextual information, and
each image comes with 5 human-annotated captions.
The images create a challenging testbed for image captioning and are widely used
in recent automatic image captioning work.

\subsection{Metrics}
The quality of generated captions is measured automatically using BLEU \cite{BLEU2002} and METEOR \cite{Meteor2014}.
BLEU roughly measures the fraction of $N$-grams (up to 4 grams) that are in common between a hypothesis and one or more references, and penalizes short hypotheses by a brevity penalty term.\footnote{We use the length of the reference that is closest to the length of the hypothesis to compute the brevity penalty.}
METEOR \cite{Meteor2014} measures unigram precision and recall, extending exact word matches to include similar words based on WordNet synonyms and stemmed tokens.
We also report the perplexity (PPLX) of studied detection-conditioned LMs.
The PPLX is in many ways the natural measure of a statistical LM, but can be loosely correlated with BLEU \cite{Auli2013}.

\subsection{Model Comparison}

\begin{table}[t]
	\centering\small
	\begin{tabular}{p{2.8cm}ccc}
		\toprule
		{\bf LM} & {\bf PPLX} & {\bf BLEU} & {\bf METEOR} \\
		\hline\rule[-1ex]{0pt}{3.5ex}%
		D-ME$^\dag$    			& 18.1 	    & 23.6 	& 22.8 \\
		\hline\rule[-1ex]{0pt}{3.5ex}%
		D-LSTM     			& 14.3 	& 22.4	& 22.6 \\
		\hline\rule[-1ex]{0pt}{3.5ex}%
		MRNN            & 13.2  & 25.7    & 22.6 \\
		\hline\rule[-1ex]{0pt}{3.5ex}%
		$k$-Nearest Neighbor  & -     & 26.0   & 22.5 \\
		1-Nearest Neighbor    & -     & 11.2   & 17.3 \\
		\bottomrule
	\end{tabular}
	\caption{Model performance on testval. \dag: From \cite{MSRCapVisCVPR2015}.}
	\label{tab:exp_lms}
\end{table}

\renewcommand{\arraystretch}{1.2}
\begin{table}
\resizebox{\columnwidth}{!}{
\begin{tabular}{lll}
\multirow{4}{0.2\textwidth}{\includegraphics[width=0.2\textwidth]{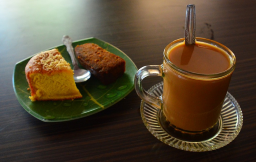}} 
& {\footnotesize D-ME+DMSM} & a plate with a sandwich and a cup of coffee \\
& {\footnotesize MRNN} & a close up of a plate of food
 \\
& {\footnotesize D-ME+DMSM+MRNN} & a plate of food and a cup of coffee  \\
& {\footnotesize $k$-NN} & a cup of coffee on a plate with a spoon \vspace{1em}\\
\multirow{4}{0.2\textwidth}{\includegraphics[width=0.2\textwidth]{figures/COCO_val2014_000000573391.png}} 
& {\footnotesize D-ME+DMSM} & a black bear walking across a lush green forest \\
& {\footnotesize MRNN} & a couple of bears walking across a dirt road \\
& {\footnotesize D-ME+DMSM+MRNN} & a black bear walking through a wooded area  \\
& {\footnotesize $k$-NN} & a black bear that is walking in the woods \vspace{1em}\\
\multirow{4}{0.2\textwidth}{\includegraphics[width=0.2\textwidth]{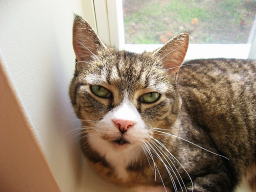}} 
& {\footnotesize D-ME+DMSM} & a gray and white cat sitting on top of it \\
& {\footnotesize MRNN} & a cat sitting in front of a mirror \\
& {\footnotesize D-ME+DMSM+MRNN} & a close up of a cat looking at the camera  \\
& {\footnotesize $k$-NN} & a cat sitting on top of a wooden table \vspace{.5em}\\
\end{tabular}
}
\caption{Example generated captions.}\label{tab:cap_ex}
\end{table}
\renewcommand{\arraystretch}{1}

In Table\,\ref{tab:exp_lms}, we summarize the generation performance of our different models. The discrete detection based models are prefixed with ``D''.  Some example generated results are show in Table \ref{tab:cap_ex}. 

We see that the detection-conditioned LSTM LM produces much lower PPLX than the detection-conditioned ME LM, but its BLEU score is no better. 
% MRNN
The MRNN has the lowest PPLX, and highest BLEU among all LMs studied in our experiments. It significantly improves BLEU by 2.1 absolutely over the D-ME LM baseline.  METEOR is similar across all three LM-based methods.

Perhaps most surprisingly, the $k$-nearest neighbor algorithm  achieves a higher BLEU score than all other models. However, as we will demonstrate in Section~\ref{sec:human}, the generated captions perform significantly better than the nearest neighbor captions in terms of human quality judgements.

\subsection{$n$-best Re-Ranking}

\begin{table}[t]
	\centering\small
	\begin{tabular}{p{4cm}ccc}
		\toprule
		{\bf Re-Ranking Features} & {\bf BLEU} & {\bf METEOR} \\
		\hline\rule[-1ex]{0pt}{3.5ex}%
		D-ME $^\dag$   & 23.6	 & 22.8 \\
	    + DMSM $^\dag$ & 25.7 	 & 23.6 \\
		+ MRNN         & 26.8    & 23.3 \\
	    + DMSM + MRNN  & {\bf 27.3}    & 23.6 \\
		\bottomrule
	\end{tabular}
	\caption{Model performance on testval after re-ranking. 
		\dag: previously reported and reconfirmed BLEU scores from \cite{MSRCapVisCVPR2015}.  +DMSM had resulted in the highest score yet reported.}
	\label{tab:exp_mert} 
\end{table}

In addition to comparing the ME-based and RNN-based LMs independently, we explore whether combining these models results in an additive improvement. To this end, we use the 500-best list from the D-ME and add a score for each hypothesis from the MRNN.\footnote{The MRNN does not produce a diverse $n$-best list.} We then re-rank the hypotheses using MERT \cite{Och2003}.  As in previous work \cite{MSRCapVisCVPR2015}, model weights were optimized to maximize BLEU score on the validation set.  We further extend this combination approach to the D-ME model with DMSM scores included during re-ranking  \cite{MSRCapVisCVPR2015}.

Results are show in Table~\ref{tab:exp_mert}. We find that combining the D-ME, DMSM, and MRNN achieves a 1.6 BLEU improvement over the D-ME+DMSM.

\subsection{Human Evaluation\label{sec:human}}

Because automatic metrics do not always correlate with human judgments \cite{callison2006re,Hodosh2013}, we also performed human evaluations using the same procedure as in \newcite{MSRCapVisCVPR2015}. Here, human judges were presented with an image, a system generated caption, and a human generated caption, and were asked which caption was ``better''.\footnote{The captions were randomized and the users were not informed which was which.} For each condition, 5 judgments were obtained for 1000 images from the testval set.

Results are shown in Table~\ref{tab:human}. The D-ME+DMSM outperforms the MRNN by 5 percentage points for the ``Better Or Equal to Human'' judgment, despite both systems achieving the same BLEU score. The $k$-Nearest Neighbor system performs  1.4 percentage points worse than the MRNN, despite achieving a slightly higher BLEU score. Finally, the combined model does not outperform the D-ME+DMSM in terms of human judgments despite a 1.6 BLEU improvement. 

Although we cannot pinpoint the exact reason for this mismatch between automated scores and human evaluation, a more detailed analysis of the difference between systems is performed in Sections~\ref{sec:lang} and \ref{sec:overlap}.

\begin{table}[h]
	\centering\small
\begin{center}
\begin{tabular}{l c@{\hskip -0.1cm}c@{\hskip -0.1cm}c}
%\hline
                   & \multicolumn{2}{c}{{\bf Human Judgements}} &  \\
     \cline{2-3}\noalign{\smallskip}
    & {\bf Better} & {\bf Better} & \\
  {\bf Approach}  &  & {\bf or Equal} & {\bf BLEU} \\
  \toprule
D-ME+DMSM             & 7.8\% & 34.0\% & 25.7 \\
MRNN                  & 8.8\% & 29.0\% & 25.7 \\
D-ME+DMSM+MRNN        & 5.7\% & 34.2\% & {\bf 27.3} \\
$k$-Nearest Neighbor  & 5.5\% & 27.6\% & 26.0\\
\bottomrule
%Human & & & & & & & \\
%\hline
\end{tabular}
\caption{Results when comparing produced captions to those written by humans, as judged by humans. These are the percent of captions judged to be ``better than'' or ``better than or equal to'' a caption written by a human.\label{tab:human}}
\vspace{-5pt}
\label{tab:human}
\end{center}
\end{table}

\section{Language Analysis}
\label{sec:lang}
%In this section, we perform a qualitative analysis of the captions produced by the D-ME+DMSM and the MRNN systems.
Examples of common mistakes we observe on the testval set are shown in Table \ref{tab:errors}.  The D-ME system has difficulty with anaphora, particularly within the phrase ``on top of it'', as shown in examples (1), (2), and (3).  This is likely due to the fact that is maintains a local context window.  In contrast, the MRNN approach tends to generate such anaphoric relationships correctly.

However, the D-ME LM maintains an explicit coverage state vector tracking which attributes have already been emitted. The MRNN {\it implicitly} maintains the full state using its recurrent layer, which sometimes results in multiple emission mistakes, where the same attribute is emitted more than once.  This is particularly evident when coordination (``and'') is present (examples (4) and (5)).

\begin{table}
\resizebox{\columnwidth}{!}{
\begin{tabular}{@{}l@{\hspace{.25em}}|@{\hspace{.5em}}p{11em}@{\hspace{.5em}}|@{\hspace{.5em}}p{12em}@{}}
& {\bf D-ME+DMSM}    & {\bf MRNN} \\\hline
1 & \footnotesize{a slice of pizza sitting on top of it} & \footnotesize{a bed with a red blanket on top of it} \\\hline
2 & \footnotesize{a black and white bird perched on top of it} & \footnotesize{a birthday cake with candles on top of it} \\\hline
3 & \footnotesize{a little boy that is brushing his teeth with a toothbrush in her mouth} & \footnotesize{a little girl brushing her teeth with a toothbrush} \\\hline
4 & \footnotesize{a large bed sitting in a bedroom} & \footnotesize{a bedroom with a bed and a bed} \\ \hline
5 & \footnotesize{a man wearing a bow tie} & \footnotesize{a man wearing a tie and a tie} \\ \hline
\end{tabular}
}
\caption{Example errors in the two basic approaches.\vspace{.5em}}\label{tab:errors}
\end{table}

\subsection{Repeated Captions\label{sec:repeated}}
All of our models produce a large number of captions seen in the training and repeated for different images in the test set, as shown in Table~\ref{tab:repetitions} (also observed by \newcite{Vinyals2015} for their LSTM-based model). There are at least two potential causes for this repetition.

First, the systems often produce generic captions such as ``a close up of a plate of food'', which may be applied to many publicly available images.   This may suggest a deeper issue in the training and evaluation of our models, which warrants more discussion in future work. Second, although the COCO dataset and evaluation server\footnote{http://mscoco.org/dataset/} has encouraged rapid progress in image captioning, there may be a lack of diversity in the data.  We also note that although caption duplication is an issue in all systems, it is a greater issue in the MRNN than the D-ME+DMSM.
\begin{table}
\centering
\small
\begin{tabular}{l|c|c}
{\bf System} & {\bf Unique} & {\bf Seen In} \\
 & {\bf Captions} & {\bf  Training} \\\hline
Human & 99.4\% & 4.8\% \\
D-ME+DMSM & 47.0\% & 30.0\% \\
MRNN & 33.1\% & 60.3\%\\
D-ME+DMSM+MRNN & 28.5\% & 61.3\%\\
$k$-Nearest Neighbor & 36.6\% & 100\%\\
\end{tabular}
\caption{Percentage unique (Unique Captions) and novel (Seen In Training) captions for testval images. For example, 28.5\% unique means 5,776 unique strings were generated for all 20,244 images.}
\label{tab:repetitions}
\end{table}

\section{Image Diversity}
\label{sec:overlap}
The strong performance of the $k$-nearest neighbor algorithm and the large number of repeated captions produced by the systems here suggest a lack of diversity in the training and test data.\footnote{This is partially an artifact of the manner in which the Microsoft COCO data set was constructed, since each image was chosen to be in one of 80 pre-defined object categories.}

We believe that one reason to work on image captioning is to be able to caption  \textit{compositionally} novel images, where the individual \textit{components} of the image may be seen in the training, but the entire composition is often not. 

In order to evaluate results for only compositionally novel images, we bin the test images based on visual overlap with the training data. For each test image, we compute the {\tt fc7} cosine similarity with each training image, and the mean value of the 50 closest images. We then compute BLEU on the 20\% least overlapping and  20\% most overlapping subsets.

Results are shown in Table~\ref{tab:overlap}. The D-ME+DMSM outperforms the $k$-nearest neighbor approach by 2.5 BLEU on the ``20\% Least'' set, even though performance on the whole set is comparable.  Additionally, the D-ME+DMSM outperforms the MRNN by 2.1 BLEU on the ``20\% Least'' set, but performs 2.1 BLEU \textit{worse} on the ``20\% Most'' set. This is evidence that D-ME+DMSM generalizes better on novel images than the MRNN; this is further supported by the relatively low percentage of captions it generates seen in the training data (Table \ref{tab:repetitions}) while still achieving reasonable captioning performance. We hypothesize that these are the main reasons for the strong human evaluation results of the D-ME+DMSM shown in Section~\ref{sec:human}.

\begin{table}[t]
	\centering\small
	\begin{tabular}{lccc}
		\toprule
		{\bf Condition}  &  \multicolumn{3}{c}{{\bf Train/Test Visual Overlap}} \\
		&  \multicolumn{3}{c}{{\bf BLEU}} \\
		     \cline{2-4}\noalign{\smallskip}
		& {\bf Whole}   & {\bf 20\%}  & {\bf 20\%} \\
        & {\bf Set}     & {\bf Least}  & {\bf Most} \\ \hline
        D-ME+DMSM             & 25.7  & 20.9 & 29.9 \\
        MRNN                  & 25.7  & 18.8 & 32.0 \\
        D-ME+DMSM+MRNN        & 27.3  & 21.7 & 32.0 \\
        $k$-Nearest Neighbor  & 26.0  & 18.4 & 33.2 \\
		\bottomrule
	\end{tabular}
	\caption{Performance for different portions of testval, based on visual overlap with the training.}
	\label{tab:overlap}
\end{table}

\section{Conclusion}
\label{sec:conclusion}
% conclusion

We have shown that a gated RNN conditioned directly on CNN activations (an MRNN) achieves better BLEU performance than an ME LM or LSTM conditioned on a set of discrete activations; and a similar BLEU performance to an ME LM combined with a DMSM. However, the ME LM + DMSM method significantly outperforms the MRNN in terms of human quality judgments. We hypothesize that this is partially due to the lack of novelty in the captions produced by the MRNN. In fact, a $k$-nearest neighbor retrieval algorithm introduced in this paper performs similarly to the MRNN in terms of both automatic metrics and human judgements.

When we use the MRNN system alongside the DMSM to provide additional scores in MERT re-ranking of the $n$-best produced by the image-conditioned ME LM, we advance by 1.6 BLEU points on the best previously published results on the COCO dataset. Unfortunately, this improvement in BLEU does not translate to improved human quality judgments.  

\newpage
\bibliographystyle{acl}
\bibliography{acl2015_imgcap}

\end{document}